\documentclass[10pt,twocolumn,letterpaper]{article}

\usepackage{iccv}
\usepackage{times}
\usepackage{epsfig}
\usepackage{graphicx}
\usepackage{amsmath}
\usepackage{amssymb}
\usepackage{kotex}
\usepackage[font=small]{caption}

\usepackage{multirow}
\usepackage{ctable}
\usepackage{tabularx,booktabs}
\newcolumntype{Y}{>{\centering\arraybackslash}X}

\usepackage[pagebackref=true,breaklinks=true,letterpaper=true,colorlinks,bookmarks=false]{hyperref} 

\iccvfinalcopy 
\def\iccvPaperID{7681}

\ificcvfinal\fi

\begin{document}
\newcommand{\Eq}[1]  {Eq.\ (#1)}
\newcommand{\Eqs}[1] {Eqs.\ (#1)}
\newcommand{\Fig}[1] {Fig.\ #1}
\newcommand{\Figs}[1]{Figs.\ #1}
\newcommand{\Tbl}[1]  {Table\ #1}
\newcommand{\Tbls}[1] {Tables\ #1}
\newcommand{\Sec}[1] {Sec.\ #1}
\newcommand{\SSec}[1] {Sec.\ #1}
\newcommand{\Secs}[1] {Secs.\ #1}
\newcommand{\Alg}[1] {Alg.\ #1}
\newcommand{\Etal}   {{\textit{et al.}}}

\newcommand{\setone}[1] {\left\{ #1 \right\}} % math set notation { a }
\newcommand{\settwo}[2] {\left\{ #1 \mid #2 \right\}} % math set notation { a | b}

\newcommand{\todo}[1]{{\textcolor{red}{#1}}}
\newcommand{\son}[1]{{\textcolor{magenta}{hyeongseok: #1}}}
\newcommand{\jy}[1]{{\textbf{\textcolor{MidnightBlue}{[JY] }}\textcolor{MidnightBlue}{#1}}}
\newcommand{\sean}[1]{{\textcolor{green}{sean: #1}}}
\newcommand{\sunghyun}[1]{{\textcolor[rgb]{0.6,0.0,0.6}{sunghyun: #1}}}
\newcommand{\kkang}[1]{{\textcolor[rgb]{0.6,0.0,0.1}{kkang: #1}}}
\newcommand{\change}[1]{{\color{red}#1}}
\newcommand{\bb}[1]{\textbf{\textit{#1}}}

% LaTeX commands to reduce the spacing above and below figures
\renewcommand{\topfraction}{0.95}
\setcounter{bottomnumber}{1}
\renewcommand{\bottomfraction}{0.95}
\setcounter{totalnumber}{3}
\renewcommand{\textfraction}{0.05}
\renewcommand{\floatpagefraction}{0.95}
\setcounter{dbltopnumber}{2}
\renewcommand{\dbltopfraction}{0.95}
\renewcommand{\dblfloatpagefraction}{0.95}

%% usually not used, instead use the ordinary latex comment %
%\newcommand{\comment}[1]{} 
%\renewcommand{\paragraph}[1]{{\vspace{0em}\noindent\textbf{#1}}}

%JY
\newcommand{\Net}[1]{#1}
\newcommand{\Loss}[1]{$\mathcal{L}_{#1}$}
\newcommand{\cm}{\checkmark}
\newcommand{\ts}{\textsuperscript}
\newcommand\oast{\stackMath\mathbin{\stackinset{c}{0ex}{c}{0ex}{\ast}{\bigcirc}}}

% %% algorithm
% \makeatletter
% \newcommand{\StatexIndent}[1][3]{%
%   \setlength\@tempdima{\algorithmicindent}%
%   \Statex\hskip\dimexpr#1\@tempdima\relax}
% \makeatother

% \newdimen{\algindent}
% \setlength\algindent{1.5em}
% \algnewcommand\LeftComment[2]{%
% \hspace{#1\algindent}$\triangleright$ \eqparbox{COMMENT}{#2} \hfill %
% }

\renewcommand{\paragraph}[1]{{\vspace{2pt}\noindent\textbf{#1}}}

\title{GAN Inversion for Out-of-Range Images with Geometric Transformations}

\author{Kyoungkook Kang\\
        POSTECH CSE\\
        {\tt\small kkang831@postech.ac.kr}
        \and
        Seongtae Kim\\
        POSTECH GSAI\\
        {\tt\small seongtae0205@postech.ac.kr}
        \and
        Sunghyun Cho\\
        POSTECH CSE \& GSAI\\
        {\tt\small s.cho@postech.ac.kr}
}

\twocolumn[{
\renewcommand\twocolumn[1][]{#1}
\maketitle
\begin{center}
    \centering
    \vspace{-0.5cm}
    \captionsetup{type=figure}
    \includegraphics[width=0.95\textwidth]{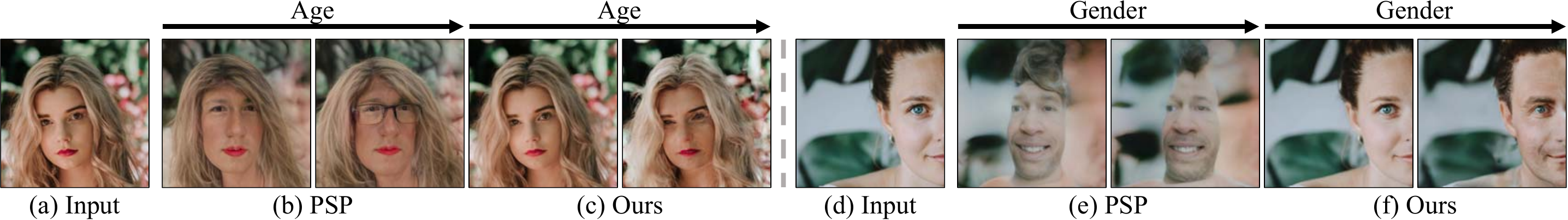}
    \vspace{-0.3cm}
    \captionof{figure}{Qualitative comparison of reconstruction and semantic editing of different methods on in-the-wild images. (a) and (d) show input images, (b) and (e) show the results of PSP~\cite{psp}, and (c) and (f) show our results.}
    \label{fig:teaser}
\end{center}
}]

\ificcvfinal\thispagestyle{empty}\fi

\begin{abstract}
\vspace{-0.3cm}
For successful semantic editing of real images, it is critical for a GAN inversion method to find an in-domain latent code that aligns with the domain of a pre-trained GAN model.
Unfortunately, such in-domain latent codes can be found only for in-range images that align with the training images of a GAN model.
In this paper, we propose \emph{BDInvert}, a novel GAN inversion approach to semantic editing of out-of-range images that are geometrically unaligned with the training images of a GAN model.
To find a latent code that is semantically editable, BDInvert inverts an input out-of-range image into an alternative latent space than the original latent space.
We also propose a regularized inversion method to find a solution that supports semantic editing in the alternative space.
Our experiments show that BDInvert effectively supports semantic editing of out-of-range images with geometric transformations.
\end{abstract}

\vspace{-0.5cm}
\section{Introduction}
\label{sec:introduction}

Generative adversarial networks (GANs) are generative models that can synthesize realistic-looking images~\cite{GAN}.
Typically, GANs learn a mapping function from a random noise vector sampled from a pre-defined distribution to a realistic-looking image through the adversarial training of a generator and a discriminator.
For the past several years, a significant progress has been made to improve the quality and diversity of synthesized images~\cite{dcgan,pggan,stylegan,stylegan2,biggan}.
As a result, recent GAN models such as StyleGAN~\cite{stylegan}, StyleGAN2~\cite{stylegan2}, and BigGAN~\cite{biggan} can produce extremely high-quality images of high resolution.

Recently, it has been shown that rich semantic information is encoded in the intermediate features and the latent space of GANs,
and furthermore, that images can be effectively edited in a semantically meaningful way by modifying features or latent code~\cite{dcgan,gan_dissection,pie,interface_gan,ganspace}.
To enable such semantic editing for real images, GAN inversion has attracted much attention lately~\cite{semantic_photo_manipulation,im2stylegan,indomain_gan,pnorm}.
GAN inversion maps a real image into the latent space of a pre-trained GAN model.
Once an inverted latent code is obtained, the image can be semantically edited by modifying its latent code or intermediate features generated from the code.

For successful semantic editing of real images, it is critical to find an \emph{in-domain} latent code that aligns with the domain of a pre-trained GAN model \cite{indomain_gan}. 
As shown in \cite{indomain_gan}, there may exist more than one latent codes that can reconstruct a given input image, and some of them may be out of the domain.
The semantic knowledge encoded in the latent space does not apply for such out-of-domain codes, thus semantic editing of such codes fails to produce proper results.

Unfortunately, such in-domain latent codes can be found only for a small fraction of real images that align with the training images of a pre-trained GAN model.
For example, most GAN models use geometrically aligned face images as their training data for ease of training.
As a result, images with a small amount of translation or other geometric transformations are out of their ranges, and the previous GAN inversion methods cannot find in-domain latent codes for such \emph{out-of-range} images.
This severely limits the applicability of semantic editing of real images using GAN inversion.
\Fig{\ref{fig:teaser}} shows real-world examples.
The input images in (a) and (d) are random images downloaded from internet.
As they are out-of-range with different rotation, scaling and translation with respect to the training dataset (FFHQ~\cite{stylegan}),
directly applying a previous GAN inversion method~\cite{psp} produces unacceptable results as shown in (b) and (e).

One solution would be to align a target image before GAN inversion, but accurate alignment of an image to the training data can be difficult or even impossible especially in the case of arbitrary natural images.
For example, for the image in \Fig{\ref{fig:teaser}}(d), a face alignment method~\cite{Kazemi14} completely fails due to severe cropping.

In this paper, we propose a novel GAN inversion approach to semantic editing of \emph{out-of-range} images, which is dubbed \emph{Base-Detail Invert (BDInvert)}.
BDInvert inverts a geometrically unaligned image with the training images for StyleGAN \cite{stylegan} and StyleGAN2 \cite{stylegan2}.
Specifically, BDInvert is designed to cover geometric transformations such as translation, rotation, and scaling, and supports various types of editing for out-of-range images that are not supported by previous approaches.

Our key idea is as follows.
It is impossible to invert an out-of-range image to an in-domain latent code in the original latent space of a pre-trained GAN model.
Instead, we propose to invert an image into another space that we refer to as the $\mathcal{F}/\mathcal{W}^+$, which consists of two subspaces $\mathcal{F}$ and $\mathcal{W}^+$.
The base code space $\mathcal{F}$ encodes geometric transformations and also supports diverse local variations that enable more faithful reconstruction of an input image.
On the other hand, the detail code space $\mathcal{W}^+$ is independent of geometric transformations and supports semantic manipulations.

To find a latent code in the $\mathcal{F}/\mathcal{W}^+$ space that faithfully reconstructs an input image, we adopt an optimization-based approach.
However, na\"{i}ve optimization of a reconstruction loss does not guarantee a latent code that supports semantic editing.
To enable semantic editing, we also propose a regularization approach based on an encoder network.
\Fig{\ref{fig:teaser}}(c) and (f) show our reconstruction and editing results of real-world images.
Thanks to our $\mathcal{F}/\mathcal{W}^+$ space and inversion approach, we can successfully reconstruct and edit the out-of-range real-world input images.

Our main contributions can be summarized as follows.
\begin{itemize}
    \vspace{-0.1cm}
    \item We propose \emph{BDInvert}, a novel GAN inversion approach to semantic editing of real images with geometric transformations that are not aligned with the training images of a pre-trained GAN model.
    \item BDInvert projects an image into an alternative latent space $\mathcal{F}/\mathcal{W}^+$ that supports more faithful reconstruction and semantic editing of out-of-range images with geometric transformations and diverse local variations.
    \item We propose a novel regularization method to find a proper solution in the $\mathcal{F}/\mathcal{W}^+$ space that supports semantic image editing.
\end{itemize}

\section{Related Work}
\label{sec:relatedwork}

In order to embed real images into the latent space of GANs, various approaches have been proposed in two directions.
One direction is to train an encoder using a data-driven approach \cite{zhu2016generative, collaborative_learning, psp}.
The other direction is to initialize a latent vector randomly or to the output of a pre-trained encoder, and then to optimize it to reconstruct a target image \cite{zhu2016generative, semantic_image_inpainting, inverting_gan, multicode_gan_inversion, semantic_photo_manipulation}.
However, inverting a real image remains a difficult problem because of the limited expressiveness of the latent space of GANs.

Recently, in order to enhance the inversion quality, several attempts to widen the latent space have been made \cite{multicode_gan_inversion,biggan_inversion_finetuning,biggan_inversion_transforming}.
Gu \etal~\cite{multicode_gan_inversion} improved the reconstruction quality by mixing features from several latent codes.
Pan \etal~\cite{biggan_inversion_finetuning} fine-tune a generator on-the-fly for more faithful reconstruction.
Huh \etal~\cite{biggan_inversion_transforming} find geometric transformation parameters to transform an image region to be more suitable for BigGAN~\cite{biggan} inversion.
Meanwhile, Abdal \etal~\cite{im2stylegan} showed high-quality embedding results for StyleGAN~\cite{stylegan} using an extended latent space $\mathcal{W}^+$.
Afterwards, many studies focusing on StyleGAN have been proposed \cite{im2stylegan++, pnorm, indomain_gan, stylegan2, pie}.
Abdal \etal~\cite{im2stylegan++} and Karras \etal~\cite{stylegan2} optimize the noise channel for more accurate embedding.
For successful image editing, embedding an image into GAN's domain is essential.
To this end, Zhu \etal~\cite{indomain_gan} train an encoder that projects an image into StyleGAN's domain, and optimize a latent code with the guidance of the encoder.
Tewari \etal~\cite{pie} introduced a hierarchical optimization that first embeds an image into the $\mathcal{W}$ space and then embeds it into the $\mathcal{W}^+$ space for better editing.
Zhu \etal~\cite{pnorm} proposed the $\mathcal{P}-norm^+$ space for in-domain inversion.
However, most existing works cannot handle out-of-range images.

\paragraph{Semantic editing}
A widely used approach to semantic image editing using GAN is to modify a latent code along semantically meaningful directions.
H\"{a}rk\"{o}nen \etal~\cite{ganspace} identify semantic directions by applying the principal component analysis (PCA) on sampled latent codes.
Shen \etal~\cite{interface_gan} use attribute classifiers to discover semantic directions.
Shen and Zhou~\cite{sefa} proposed an unsupervised method that factorizes the weights of latent code transformation layers to find semantic directions that cause large changes to the output. 
\section{Latent Space $\mathcal{F}/\mathcal{W}^+$}
\label{sec:latent_space}

\begin{figure*}[t]
\centering
\includegraphics[width=0.91\linewidth]{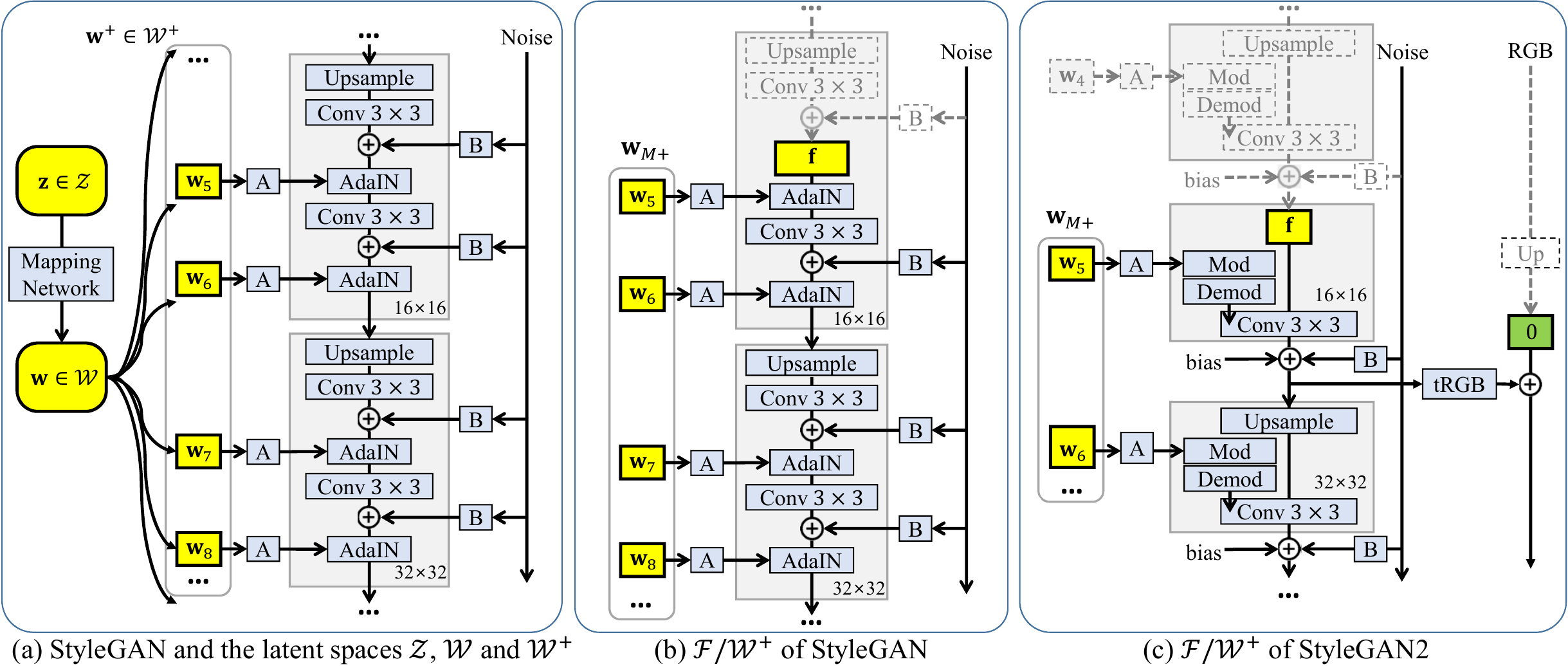}
\vspace{-0.2cm}
\caption{The network architectures of StyleGAN~\cite{stylegan} and StyleGAN2~\cite{stylegan2} and the layers corresponding to their latent spaces marked in yellow.
The layers that are not used for the latent space $\mathcal{F}/\mathcal{W}^+$ are marked with gray dotted borders in (b) and (c).
In StyleGAN2, the RGB image layer from the coarse scale is replaced with a tensor filled with zero as indicated by a green box.}
\label{fig:latent_space}
\vspace{-0.3cm}
\end{figure*}

In this section, we first review state-of-the-art GAN inversion approaches and discuss their limitations on out-of-range images.
Then, we introduce an alternative latent space $\mathcal{F}/\mathcal{W}^+$ to overcome the limitations.

Our approach is based on StyleGAN and StyleGAN2~\cite{stylegan,stylegan2}, which produce high-quality synthesis results.
Both GAN frameworks use a mapping network $f:\mathcal{Z}\rightarrow\mathcal{W}$ based on a multi-layer perceptron (MLP) that maps a latent code $\mathbf{z} \in \mathcal{Z}$ to an intermediate latent code $\mathbf{w} \in \mathcal{W}$ as shown in \Fig{\ref{fig:latent_space}}(a).
Compared to the latent space $\mathcal{Z}$, the intermediate latent space $\mathcal{W}$ provides less entangled representations of different attributes so that different attributes can be more easily adjusted in the image generation process.
Another noticeable feature of StyleGAN and StyleGAN2 is their multi-scale image synthesis approaches, which enable scale-wise disentanglement of different attributes.
To control the generation process in a multi-scale manner, both StyleGAN and StyleGAN2 feed the intermediate latent code $\mathbf{w}$ to multiple layers of different scales of the generator.
In addition, to enhance the diversity of synthesized images, both StyleGAN and StyleGAN2 utilize noise randomly sampled from a Gaussian distribution for each image generation.

While $\mathcal{W}$ is effective in generating diverse images with different attributes,
it is still not sufficient for GAN inversion of a wide range of real images.
To enhance the reconstruction accuracy, Abdal \etal~\cite{im2stylegan} proposed an extended latent space $\mathcal{W}^+$.
Each element $\mathbf{w}^+ \in \mathcal{W}^+$ is defined as $\mathbf{w}^+ = \{\mathbf{w}_1,\mathbf{w}_2, \cdots, \mathbf{w}_N\}$ where $\mathbf{w}_i$ is a latent code in $\mathcal{W}$ and $N$ is the number of layers in the generator that takes $\mathbf{w}$ as input (\Fig{\ref{fig:latent_space}}(a)).
The subscript $i$ in $\mathbf{w}_i$ is the index of a layer that takes $\mathbf{w}$ such that $i=1$ and $i=N$ indicate the first and last layers in the smallest and largest scales, respectively.
With the extended latent space $\mathcal{W}^+$, different latent codes can be used for different layers, and consequently, a wider range of images can be reconstructed.

Later, Zhu \etal~\cite{indomain_gan} showed that, for semantic image manipulation, it is essential to find an \emph{in-domain} latent code instead of a latent code that precisely reconstructs an input image.
They also showed that real images can be effectively inverted to an in-domain latent code in $\mathcal{W}^+$ with a domain-guided encoder and domain-regularized optimization.

Nonetheless, GAN inversion to the extended latent space $\mathcal{W}^+$ still fails to find an in-domain latent code for \emph{out-of-range} images as discussed in \Sec{\ref{sec:introduction}}.
To overcome this limitation, we propose another latent space $\mathcal{F}/\mathcal{W}^+$.
Each element $\mathbf{w}^*$ in $\mathcal{F}/\mathcal{W}^+$ is defined as $\mathbf{w}^*=(\mathbf{f}, \mathbf{w}_{M+})$ where $\mathbf{f}$ is a base code and $\mathbf{w}_{M+}$ is a detail code. 
$\mathbf{w}_{M+}$ is a set of latent codes for the fine scales of the generator, which is defined as $\mathbf{w}_{M+}=\{\mathbf{w}_{M}, \cdots, \mathbf{w}_N\}$.
$\mathbf{f}$ is a coarse-scale feature map of the generator before the layer that takes $\mathbf{w}_M$.
Specifically, for StyleGAN~\cite{stylegan}, we define $\mathbf{f}$ as the feature map right before the first adaptive instance normalization (AdaIN) layer~\cite{adain} at a certain scale.
For StyleGAN2~\cite{stylegan2}, we define $\mathbf{f}$ as the feature map after a pair of upsampling and convolution layers at a certain scale.
\Fig{\ref{fig:latent_space}}(b) and (c) depict the latent space $\mathcal{F}/\mathcal{W}^+$ of StyleGAN and StyleGAN2, respectively.
In our experiments, we test two different scales, $8\times8$ and $16\times16$, for $\mathbf{f}$.

In the case of StyleGAN2~\cite{stylegan2}, the generator needs a feature map corresponding to an RGB image upsampled from the previous scale (\Fig{\ref{fig:latent_space}}(c)).
While we may include a small-scale feature map as a part of our latent space, we observed that the feature maps at the coarse scales have values close to zero and have little impact on image generation results.
Thus, we simply set them to zero in our experiments as depicted by the green box in \Fig{\ref{fig:latent_space}}(c).

The $\mathcal{F}/\mathcal{W}^+$ space provides a couple of nice properties that enable semantic editing of out-of-range images.
First, compared to $\{\mathbf{w}_1, \cdots, \mathbf{w}_{M-1}\}$, the base code $\mathbf{f}$ can represent a wider range of images including images with geometric transformations.
For example, as $\mathbf{f}$ is a feature map of a convolutional neural network (CNN), we can simply shift $\mathbf{f}$ along the $x$- or $y$-axis to represent the feature map of a shifted image.
Second, the detail code $\mathbf{w}_{M+}$ is invariant to translations of images.
Specifically, in the case of StyleGAN~\cite{stylegan}, $\mathbf{w}_{M+}$ controls the parameters of the AdaIN~\cite{adain} layers of the generator.
Similarly, in the case of StyleGAN2~\cite{stylegan2}, $\mathbf{w}_{M+}$ controls the parameters of the demodulation layers.
Both AdaIN and demodulation operations are global operations that are applied to CNN features in a translation-invariant manner.

\begin{figure}[t]
\centering
\includegraphics[width=0.95\linewidth]{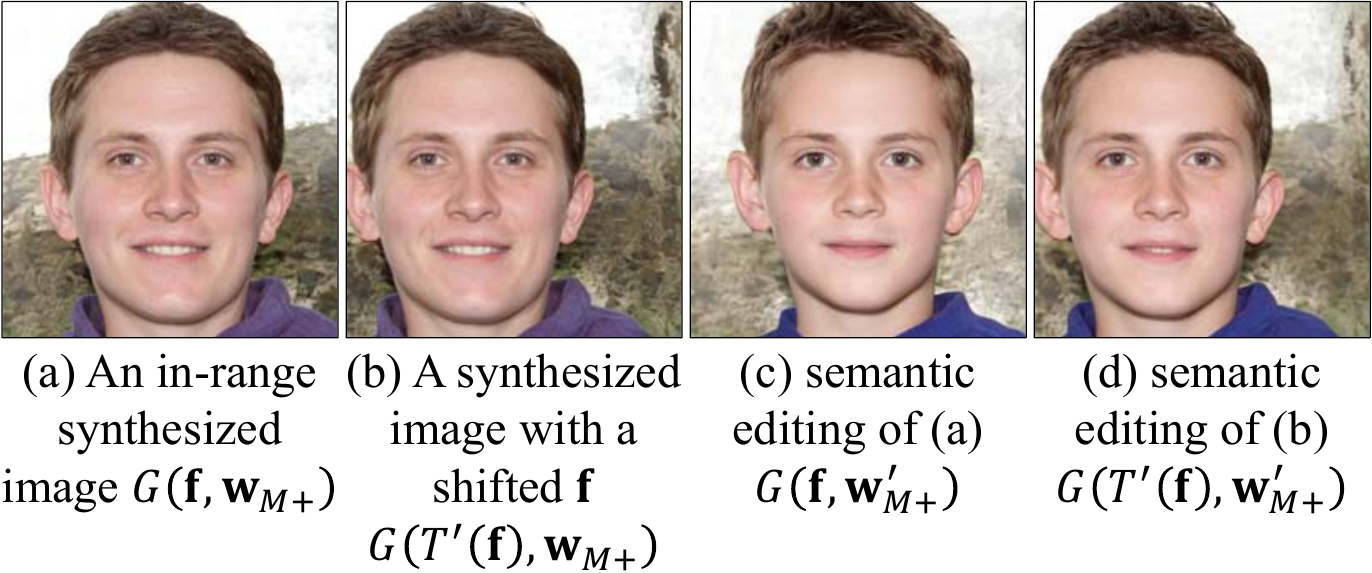}
\vspace{-0.2cm}
\caption{Semantic editing in the $\mathcal{F}/\mathcal{W}^+$ space. (a) An in-range synthesized image generated from an in-domain latent code $(\mathbf{f},\mathbf{w}_{M+})$. (b) By applying a geometric transform to $\mathbf{f}$, an out-of-range image can be obtained. Because $\mathbf{w}_{M+}'$ affects the image globally, an image editing operation used for an in-range image in (c) can be used for an out-of-range image in (d).}
\label{fig:image_synthesis_with_shifted_f}
\vspace{-0.3cm}
\end{figure}
Thanks to the aforementioned properties, we can describe the relationship between an image $I$ and its transformed image $T(I)$ where $T$ is a geometric transformation operator as follows.
Suppose that $I$ is generated from $\mathbf{w}^*$, i.e., $I=G(\mathbf{w}^*)=G(\mathbf{f},\mathbf{w}_{M+})$
where $G$ is the generator of a pre-trained GAN model.
Then, $T(I)$ can be expressed as:
\begin{equation}
   T(I) \approx G(T'(\mathbf{f}), \mathbf{w}_{M+})
   \label{eq:image_model}
\end{equation}
where $T'$ is a geometric transformation operator corresponding to $T$ whose scale is adjusted according to the relative scale of $\mathbf{f}$ to $I$.
This relationship can be also used for semantic image manipulation of $T(I)$.
As $T'(\mathbf{f})$ is a CNN feature map and $\mathbf{w}_{M+}$ is a set of parameters for global operations,
for editing $T(I)$, we can manipulate $\mathbf{w}_{M+}$ in the same way for $I$ and achieve similar editing results.

\Fig{\ref{fig:image_synthesis_with_shifted_f}} shows an example that illustrates the relationship in \Eq{\ref{eq:image_model}}.
In this example, we sample an in-domain latent code $(\mathbf{f},\mathbf{w}_{M+})$ and generate an in-range image in \Fig{\ref{fig:image_synthesis_with_shifted_f}}(a) using StyleGAN2~\cite{stylegan2}.
Shifting $\mathbf{f}$, we can generate a shifted image of \Fig{\ref{fig:image_synthesis_with_shifted_f}}(a) as shown in \Fig{\ref{fig:image_synthesis_with_shifted_f}}(b).
While they are not exactly the same due to the zero padding and noise component in StyleGAN2, they look almost identical proving the relationship in \Eq{\ref{eq:image_model}}.
\Fig{\ref{fig:image_synthesis_with_shifted_f}}(c) and (d) show the semantic editing results of (a) and (b) using the same manipulated latent code $\mathbf{w}'_{M+}$.
The results show that we can effectively perform semantic editing for geometrically transformed images in the same way as for in-range images.

The discussion above shows that, as long as $(\mathbf{f},\mathbf{w}_{M+})$ is in-domain, $(T'(\mathbf{f}),\mathbf{w}_{M+})$ for an arbitrary $T'$ also supports semantic image editing.
Based on this, we define an extended domain of $\mathbf{w}^*$ as a set of geometrically transformed latent codes $(T'(\mathbf{f}),\mathbf{w}_{M+})$ of in-domain latent codes $(\mathbf{f},\mathbf{w}_{M+})$ for arbitrary transformations $T'$.

While the discussion above discusses only geometric transformations, we note that our latent space $\mathcal{F}/\mathcal{W}^+$ supports not only geometric transformations but also diverse local variations as the base code $\mathbf{f}$ supports locally different information.
This leads to more faithful reconstruction even for images without geometric transformations as will be shown in \Sec{\ref{sec:experiments}}.
We also note that the latent space $\mathcal{F}/\mathcal{W}^+$ does not support semantic editing that require coarse-scale $\mathbf{w}_i$'s such that $i < M$.
However, our experiments show that it still supports various types of semantic editing as we define $\mathbf{f}$ as a very coarse-scale feature map.

\section{Regularized Inversion to $\mathcal{F}/\mathcal{W}^+$}

For inversion of an image, we adopt the optimization-based approach since it generally achieves higher reconstruction quality compared to the encoder-based approach~\cite{inverting_gan,multicode_gan_inversion,indomain_gan}.
In this section, we introduce our optimization approach both for StyleGAN and StyleGAN2~\cite{stylegan,stylegan2}.

\subsection{Reconstruction Loss}
Given an input image $I$, to find a latent code $\textbf{w}^*$ that reconstructs $I$,
we optimize an objective function with a reconstruction loss $L_{recon}$, which is defined as:
\begin{equation}
    L_{recon}(\textbf{w}^*) = L_{MSE}(\textbf{w}^*) + \omega_{per} L_{per}(\textbf{w}^*)
    \label{eq:L_recon}
\end{equation}
where $L_{MSE}$ and $L_{per}$ are mean-squared-error (MSE) and perceptual losses, respectively.
$\omega_{per}$ is a weight for $L_{per}$.
$L_{MSE}$ is defined as
%\begin{equation}
    $L_{MSE}(\textbf{w}^*) = \| I - G(\textbf{w}^*) \|^2$
%\end{equation}
where $G$ is the generator of a pre-trained StyleGAN model.
$L_{per}$ is defined as
%\begin{equation}
    $L_{per}(\textbf{w}^*) = \| F(I) - F(G(\textbf{w}^*)) \|^2$,
%\end{equation}
where $F$ is a LPIPS network to compute the perceptual distance~\cite{lpips}.

\begin{figure}[t]
\centering
\includegraphics[width=0.95\linewidth]{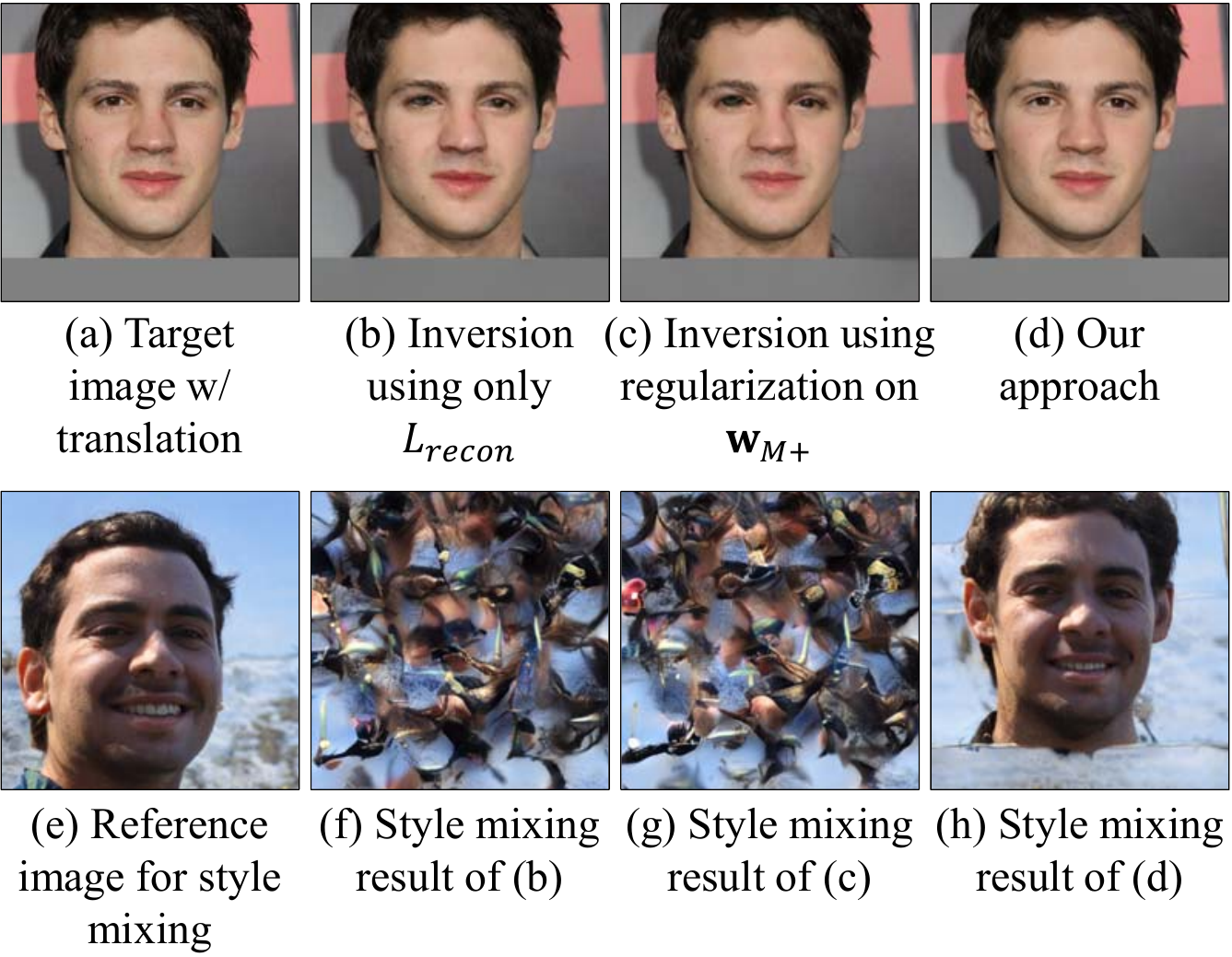}
\vspace{-0.2cm}
\caption{Inversion to $\mathcal{F}/\mathcal{W}^+$ with different combinations of the loss terms. The results of the style mixing~\cite{stylegan} operation in (f), (g) and (h) are obtained by replacing their $\mathbf{w}_{M+}$'s by $\mathbf{w}_{M+}$ from the reference image in (e).}
\label{fig:inversion_examples}
\vspace{-0.3cm}
\end{figure}

By optimizing \Eq{\ref{eq:L_recon}}, e.g., using the Adam optimizer~\cite{adam}, we can obtain latent codes that produce high-quality reconstruction results even for out-of-range images thanks to the high expressive power of the $\mathcal{F}/\mathcal{W}^+$ space.
However, such latent codes do not support semantic image editing as they are out-of-domain.
\Fig{\ref{fig:inversion_examples}} shows an example using StyleGAN2~\cite{stylegan2}.
\Fig{\ref{fig:inversion_examples}}(a) shows a target image, which is out-of-range due to translation.
\Fig{\ref{fig:inversion_examples}}(e) is a reference image for style mixing, which is a semantic image editing operation~\cite{stylegan}.
Optimizing the reconstruction loss in \Eq{\ref{eq:L_recon}}, we can obtain a latent code that accurately reconstructs the target image as shown in \Fig{\ref{fig:inversion_examples}}(b).
However, the estimated latent code is out-of-domain, so it fails to produce an appropriate style mixing result as shown in \Fig{\ref{fig:inversion_examples}}(f).

To enable semantic editing of out-of-range images, both $\mathbf{f}$ and $\mathbf{w}_{M+}$ must be in proper domains. 
To guide our optimization process to a solution in a proper domain,
we adopt regularization both on $\mathbf{f}$ and $\mathbf{w}_{M+}$.
The following subsections discuss our regularization schemes one by one.

\subsection{Regularization on Detail Code $\mathbf{w}_{M+}$}

To promote in-domain $\mathbf{w}_{M+}$,
we adopt the $\mathcal{P}-norm^+$ space-based regularization scheme proposed by Zhu \etal~\cite{pnorm}.
Specifically, at each iteration of the iterative optimization of our objective function,
we transform the current estimate of $\mathbf{w}_{M+}$ into the $\mathcal{P}-norm^+$ space. 
Then, we clip the values that are out of a certain range.
In our experiments, we used the range $[-5\sigma,5\sigma]$ as suggested in \cite{pnorm} where $\sigma$ is the standard deviation of in-domain latent codes.
We then transform the clipped values back into the $\mathcal{W}^+$ space.
We refer the readers to \cite{pnorm} for more details.

\subsection{Regularization on Base Code $\mathbf{f}$}
While optimizing \Eq{\ref{eq:L_recon}} with the regularization on $\mathbf{w}_{M+}$ results in an in-domain solution for $\mathbf{w}_{M+}$, it still produces an improper solution for $\mathbf{f}$ that results in the failure of semantic image editing.
\Fig{\ref{fig:inversion_examples}}(c) shows an inversion result using the reconstruction loss with the regularization on $\mathbf{w}_{M+}$.
Thanks to the hard clipping in the $\mathcal{P}-norm^+$ space, the estimated $\mathbf{w}_{M+}$ is always in a desired range.
However, the estimated $\mathbf{f}$ is still out-of-domain, and produces an incorrect style mixing result in \Fig{\ref{fig:inversion_examples}}(g).

To overcome this, we introduce a regularization method that encourages $\mathbf{f}$ to be in the extended domain of $\mathbf{f}$ defined in \Sec{\ref{sec:latent_space}}.
Our method is a two-step approach.
For an input image $I$, we first find an initial base code $\mathbf{f}^o$ that lies in the extended domain of $\mathbf{f}$ using an encoder $E$.
Then, while optimizing \Eq{\ref{eq:L_recon}}, we find a base code $\mathbf{f}$ that is close to $\mathbf{f}^o$.
To achieve this, we define a regularization loss for $\mathbf{f}$ as:
\begin{equation}
  L_{\mathbf{f}}(\mathbf{w}^*) = \| \mathbf{f}^o - \mathbf{f} \|^2    
  \label{eq:L_f}
\end{equation}
where $\mathbf{f}^o = E(I)$. 

Our final objective function is then defined as:
\begin{equation}
    L(\mathbf{w}^*) = L_{recon}(\mathbf{w}^*) + \omega_{\textbf{f}} L_{\mathbf{f}}(\mathbf{w}^*)
    \label{eq:final_objective}
\end{equation}
where $\omega_{\mathbf{f}}$ is a weight for the regularization loss $L_{\mathbf{f}}$.
Our final approach optimizes \Eq{\ref{eq:final_objective}} with the regularization on $\mathbf{w}_{M+}$.
\Fig{\ref{fig:inversion_examples}}(d) and (h) show that our final approach can successfully invert an out-of-range image and support semantic image editing, respectively.

\subsection{Encoder for Base Code $\mathbf{f}$}
\label{sec:encoder}

Our encoder estimates an initial base code $\mathbf{f}^o$ of an input image.
As $\mathbf{f}^o$ has a small spatial resolution, e.g., $16\times16$,
the encoder does not require an input image of the original resolution or a heavy network architecture.
Thus, the encoder is designed to take a downsampled image of the resolution $8\times$ larger than $\mathbf{f}$, e.g., $128\times128$.
The encoder has a VGG-like architecture~\cite{simonyan2015vgg} consisting of 11 convolution blocks and three pooling layers without fully connected layers. 
More details can be found in the supplementary material.

For the training of the encoder, we randomly sample a batch of latent codes from the latent space $\mathcal{Z}$ at each iteration. 
From each sampled latent code $\mathbf{z}$, we obtain its corresponding latent code $(\mathbf{f}^{gt},\mathbf{w}_{M+}^{gt})$ and its image $I$.
Using the sampled latent codes and their images, we train our encoder with a loss function defined as:
\begin{eqnarray}
    L_{enc} &=& \left\| G(E(I_\downarrow),\mathbf{w}_{M+}^{gt}) - I \right\|^2 \label{eq:encoder_loss}\\
            &+& \lambda_{per} \left\| F(G(E(I_\downarrow),\mathbf{w}_{M+}^{gt}))-F(I)\right\|^2 \nonumber
\end{eqnarray}
where $I_\downarrow$ is a downsampled version of $I$.
The first and second terms on the right-hand side are a MSE loss and a perceptual loss.
The loss minimizes the difference between the training image $I$ and its reconstructed image using the latent code obtained by the encoder.

As we have $\mathbf{f}^{gt}$, we may use a loss term based on the distance between $E(I_\downarrow)$ and $\mathbf{f}^{gt}$, e.g, $\| E(I_\downarrow)-\mathbf{f}^{gt}\|^2$.
However, we found that using it instead of the loss terms in \Eq{\ref{eq:encoder_loss}} leads to less accurate reconstruction of an input image.

Our training procedure does not use geometrically transformed images.
Nevertheless, our encoder still performs effectively for geometrically transformed images thanks to the spatially-invariant property of CNNs.
For example, for a shifted image, our encoder estimates a shifted feature map $\mathbf{f}^o$ that lies in the extended domain of $\mathbf{f}$.

Although \Eq{\ref{eq:encoder_loss}} does not have any terms to encourage to predict a latent code in the extended domain,
our encoder can effectively find a latent code that supports semantic image editing. 
As the encoder is trained using a large amount of images with a large batch size, we found that it is not necessary to include any other constraints such as the loss term based on the latent code distance.

\section{Experiments}
\label{sec:experiments}

\begin{figure*}[!t]
\centering
\includegraphics[width=0.91\linewidth]{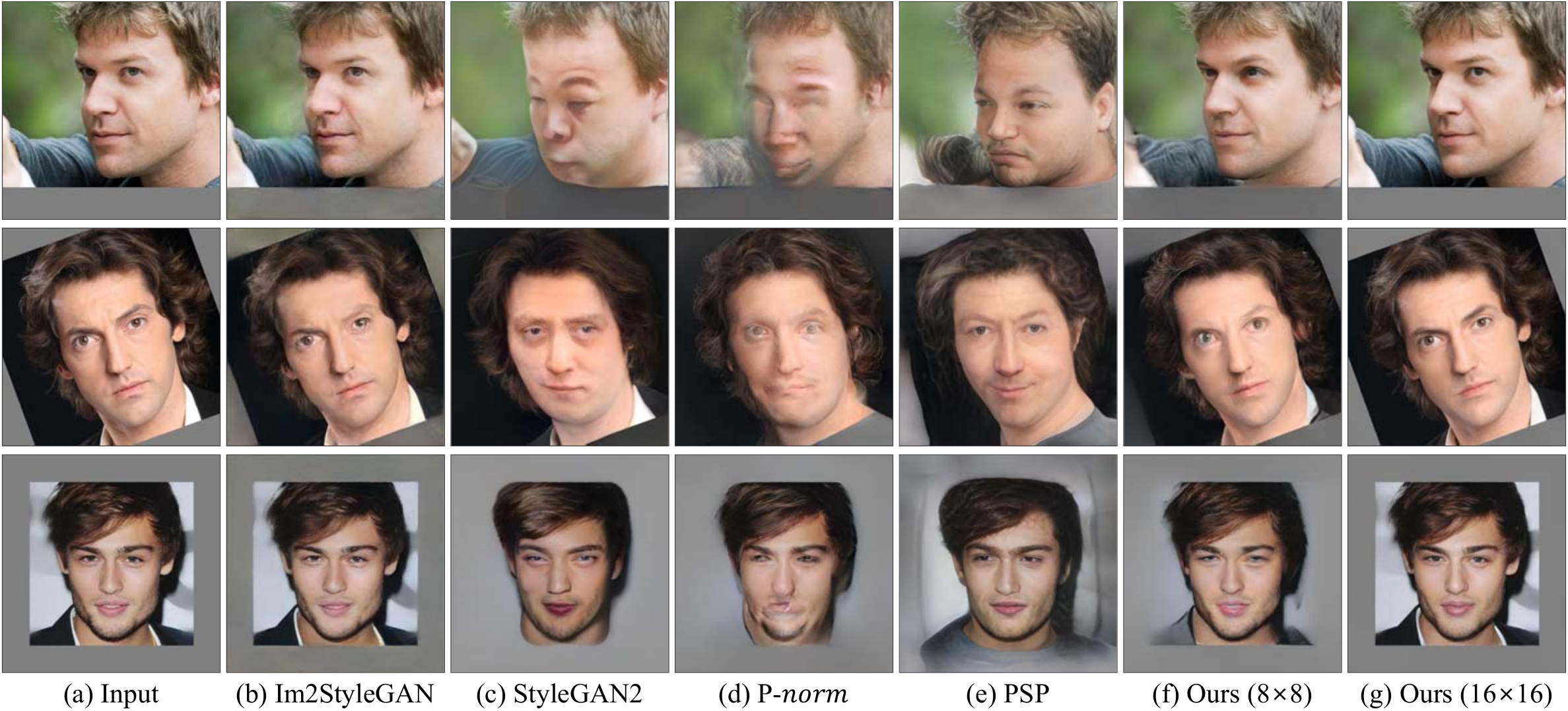}
\vspace{-0.2cm}
\caption{Qualitative comparison of the reconstruction quality of different methods. The input images are sampled from the CelebA-HQ dataset and applied different geometric transformations. Top to bottom: translation by 150 pix., rotation by 20 deg., and scaling by 3/4.}
\label{fig:ffhq_results}
\vspace{-0.1cm}
\end{figure*}

\begin{table*}[t]
    \centering
\scalebox{0.8}{
\begin{tabular}{ccc|cccc|ccc|cccc}
\specialrule{.1em}{.05em}{.05em} 
                             &   &     & \multicolumn{4}{c}{Translation} & \multicolumn{3}{c}{Rotation} & \multicolumn{4}{c}{Scaling}         \\
\hline
Models                       &  Metric & & 0   & 50   & 100   & 150  & 10       & 20      & 30      &7/8  {$\downarrow$}  &  3/4 {$\downarrow$} &  9/8 {$\uparrow$} &  5/4 {$\uparrow$} \\
\specialrule{.1em}{.05em}{.05em} 
\multirow{2}{*}{Im2StyleGAN \cite{im2stylegan}} & PSNR &   $\uparrow$& \underline{25.63} & \underline{25.06} &  \underline{24.53} & \underline{23.92} & \underline{25.76}  &  \underline{24.65}     &      \underline{23.87}    &    \underline{25.82}   &  \underline{25.25}      &   \underline{26.17}   &   \underline{26.27}  \\
                             & FID  &   $\downarrow$ &  48.37   &  \underline{45.73}    &  52.52     & 58.64     & \underline{50.06}    &  \underline{56.63}     &   \underline{65.76}     &   \underline{33.80}     &  \underline{34.24}      &  \underline{38.02}    &   \underline{36.78}   \\
\hline
\multirow{2}{*}{P-norm$^+$ \cite{pnorm}}      & PSNR &$\uparrow$  &   21.79  &  20.94    &   19.78    &  18.54    &      20.70    &  18.91       &   17.93      &    21.53     &     19.41    &      22.07 &     21.85  \\
                              &  FID & $\downarrow$  & 58.69    &  64.52    &  78.56     &  98.53    &  77.93        & 86.16        &  110.48       &   46.89      &    60.38     & 52.76      &    49.06   \\
\hline                             
\multirow{2}{*}{StyleGAN2 inv.~\cite{stylegan2}}  & PSNR &$\uparrow$  &   18.73  &  18.29  &  17.31     &     16.71 &     17.95     &     17.22    &      16.02   &  18.65       &    18.43     &     19.12  &    19.43   \\
                              & FID & $\downarrow$ &  65.49   & 70.36     & 78.32  &  87.70   &     79.31     &     82.25    &  96.23       &  52.26       &    50.23     &     60.64  &    60.24   \\
\hline                             
\multirow{2}{*}{PSP \cite{psp}}  & PSNR & $\uparrow$ &   20.54  &  19.03    &  17.59     & 16.50 &     19.14     &     17.78    &      16.99   &       19.02  &    17.78     &     20.63  &    20.15   \\
                              & FID & $\downarrow$ &  78.53   &     84.85 &     99.66  &    118.50  &       108.13   &      115.46   &      142.09   &  84.87       &  96.29       &    70.16   &  68.32     \\
\hline                             
\multirow{2}{*}{Ours ($8\times8$)}   & PSNR & $\uparrow$ &    23.69 &     23.35 &     23.74  &    23.50  &    23.30      &    22.06     &     21.35    &      23.37   &       22.72  &    23.93   &  24.22     \\
                              & FID & $\downarrow$ &  49.68   & 49.47     &    \underline{46.05}  &   \underline{49.00}  &    60.84      &    60.52     &     71.71    &      37.51   & 38.34        &    44.11   &  37.43     \\
\hline                            
\multirow{2}{*}{Ours ($16\times16$)}  & PSNR & $\uparrow$ & \textbf{26.47}   &    \textbf{26.30}  &   \textbf{26.37}   &  \textbf{26.43}    &  \textbf{26.48}   &   \textbf{26.49}     &    \textbf{26.33}     &   \textbf{26.44}    &      \textbf{26.28}   & \textbf{26.98}    &    \textbf{27.26} \\
                              &  FID & $\downarrow$ & \textbf{30.27}    &   \textbf{32.16}  &   \textbf{30.68}   & \textbf{31.58}    &  \textbf{37.01}       &       \textbf{33.96}  &   \textbf{33.98}     &   \textbf{24.92}    & \textbf{24.29}       &  \textbf{27.61}     &  \textbf{23.84}     \\
\specialrule{.1em}{.05em}{.05em}                              
\end{tabular}
}
    \vspace{-0.2cm}
    \caption{Quantitative comparison of the reconstruction quality of different methods on geometrically transformed images. For the evaluation, we sample 50 images from the CelebA-HQ dataset \cite{pggan} and applied different degrees of translation, rotation and scaling.} 
    \label{tbl:quantitative_results_on_ffhq}
    \vspace{-0.3cm}
\end{table*}

\begin{figure*}[t]
\centering
\includegraphics[width=0.85\linewidth]{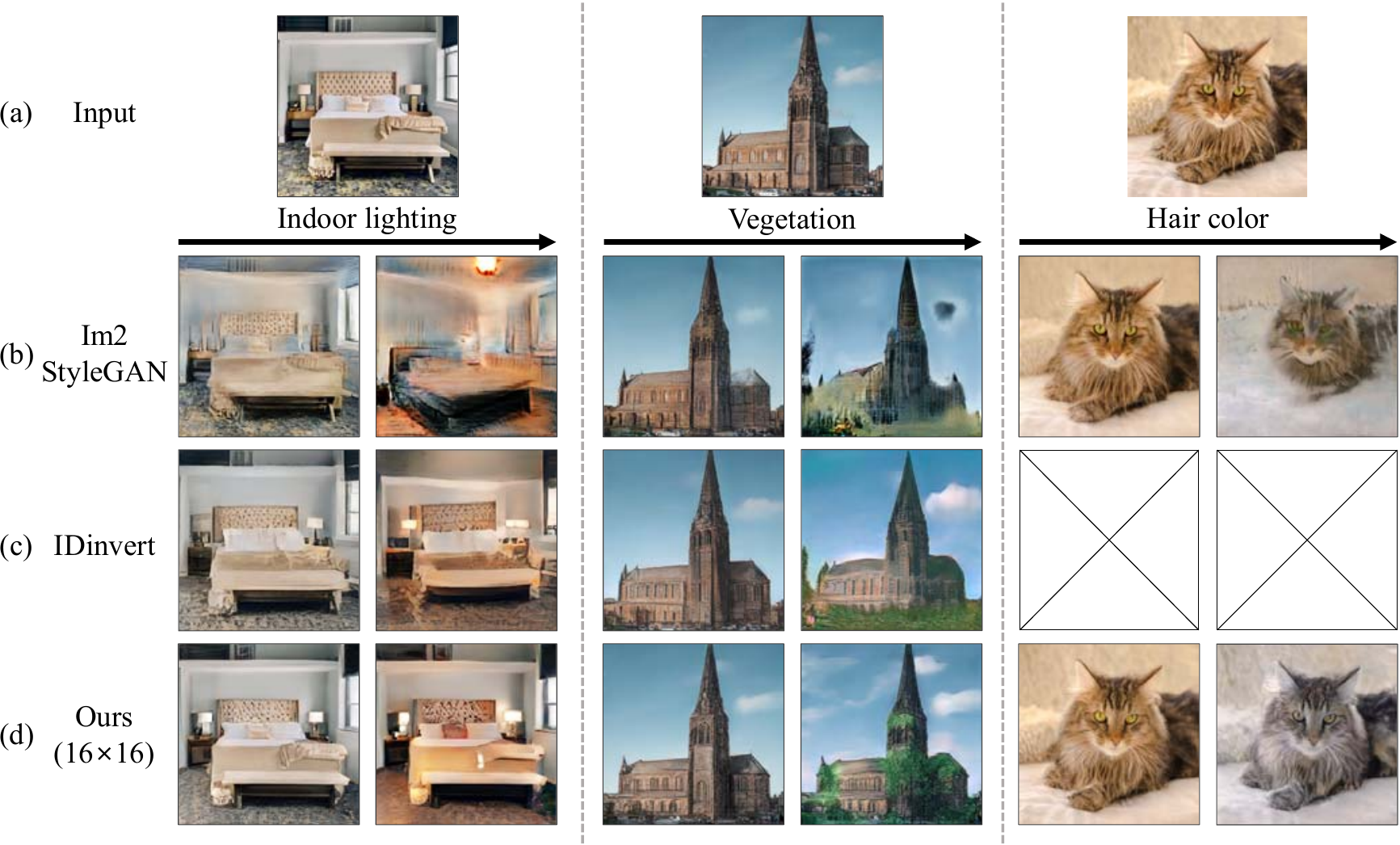}
\vspace{-0.4cm}
\caption{
Qualitative comparison of the reconstruction and semantic editing quality of different methods on natural images. The input images on the top row are collected from the internet. For the bedroom and tower image on the left and middle, we use StyleGAN~\cite{stylegan} models pre-trained on the LSUN bedroom and tower datasets~\cite{lsun}. For the cat image on the right, we use a StyleGAN2~\cite{stylegan2} model pre-trained on the LSUN cat dataset. For the cat dataset, the results of IDinvert~\cite{indomain_gan} are not available as IDinvert does not provide pre-trained weights for its encoder network. From left to right, the semantic editing operations are indoor lighting, vegetation and hair color change.
}
\label{fig:general_results}
\vspace{-0.3cm}
\end{figure*}

\begin{table}[t]
    \centering
\scalebox{0.8}{
\begin{tabular}{ccc|c|c|c}
\specialrule{.1em}{.05em}{.05em} 
                             &   &     & \multicolumn{3}{c}{Dataset}       \\
\hline
Models                       &  Metric & & Bedroom   & Tower   & Cat    \\
\specialrule{.1em}{.05em}{.05em} 
\multirow{2}{*}{Im2StyleGAN} & PSNR &   $\uparrow$& \underline{19.88} & \textbf{20.64} & \underline{22.90}    \\
                             & FID  &   $\downarrow$ &  111.73   &  \underline{58.14}    & \underline{71.19}     \\
\hline
\multirow{2}{*}{IDinvert }      & PSNR &$\uparrow$  &  19.27  & 20.02    &  -    \\
                              &  FID & $\downarrow$  & 80.21   &  75.59    &  -     \\
\hline                            
\multirow{2}{*}{Ours (16x16)}  & PSNR & $\uparrow$ & \textbf{20.21}   &   \underline{20.37}  &  \textbf{24.67}  \\
                              &  FID & $\downarrow$ & \textbf{49.92}    &  \textbf{42.89}  &  \textbf{31.74}  \\
\specialrule{.1em}{.05em}{.05em}                              
\end{tabular}
}
    \vspace{-0.2cm}
    \caption{Quantitative comparison of the reconstruction quality of different methods on natural images. Each test set consists 25 images collected from the internet. For the bedroom and tower test sets, we use StyleGAN~\cite{stylegan} models pre-trained on the LSUN bedroom and tower datasets~\cite{lsun}. For the cat test set, we use a StyleGAN2~\cite{stylegan2} model pre-trained on the LSUN cat dataset. For the cat dataset, the results of IDinvert~\cite{indomain_gan} are not available as IDinvert does not provide pre-trained weights for its encoder network.}
    \label{tbl:quantitative_results_on_general_datasets}
    \vspace{-0.6cm}
\end{table}

\paragraph{Implementation details}
In our implementation, we downsample images to $256\times 256$ to compute the perceptual losses in $L$ and $L_{enc}$ following previous works \cite{im2stylegan, pulse, pnorm}.
In our experiments, we set $\omega_{per}=10$, $\omega_{\mathbf{f}}=10$ and $\lambda_{per}=10$. 
For training the encoder, we set the batch size to $16$ and the number of iterations to 10,000.
We initially set the learning rate to 0.001 and reduced it by a factor of 0.1 every 2,000 iterations.
For the inversion, we use 1,200 iterations with learning rate of 0.01.
We use the Adam optimizer~\cite{adam} both for the training of the encoder and GAN inversion.
We conducted our experiments using pre-trained models of StyleGAN\footnote{\url{https://github.com/genforce/idinvert_pytorch}} and StyleGAN2\footnote{\url{https://github.com/genforce/genforce}}.

In our experiments, we implement semantic editing operations by adding a semantic editing vector to a latent code, i.e., $\mathbf{w}^{edit} = \mathbf{w} + \alpha\mathbf{v}$ where $\alpha$ is a user parameter to control the editing strength and $\mathbf{v}$ is an editing vector following \cite{interface_gan, sefa, indomain_gan}.
Specifically, we use editing vectors provided by IDinvert \cite{indomain_gan} and SeFa \cite{sefa} for StyleGAN~\cite{stylegan} and StyleGAN2~\cite{stylegan2}, respectively.
For a latent code in $\mathcal{F}/\mathcal{W}^+$, we add an editing vector only to a detail code $\mathbf{w}_{M+}$.

\paragraph{Reconstruction comparison}
We first compare the reconstruction quality of our method with those of previous state-of-the-art inversion methods on the CelebA-HQ dataset~\cite{stylegan} using a StyleGAN2~\cite{stylegan2} model pre-trained on the FFHQ dataset~\cite{stylegan}.
For the comparison, we constructed a test set composed of 50 images randomly extracted from the CelebA-HQ dataset \cite{pggan}.
In order to investigate the inversion performance on out-of-range images with geometric transformations, we applied different transformations to the test set.
Specifically, we applied translation of 50, 100, and 150 pixels in random directions, rotation by 10, 20, and 30 degrees randomly in a counterclockwise and clockwise direction, and scaling by 7/8, 3/4, 9/8, and 5/4.

We compare our method with state-of-the-art methods: Im2StyleGAN~\cite{im2stylegan}, StyleGAN2 inversion~\cite{stylegan2}, P-norm$^+$~\cite{pnorm}, and PSP~\cite{psp}.
PSP is an encoder-based method while the others are optimization-based ones.
We used the authors' code for StyleGAN2 and PSP.
We implemented Im2StyleGAN and P-norm$^+$ as their code is not available.
We also compare two versions of our method, which use a base code $\textbf{f}$ of size $8\times 8$ and $16\times16$, respectively.

\Fig{\ref{fig:ffhq_results}} shows a qualitative comparison.
As shown in the figure, all the methods except for Im2StyleGAN~\cite{im2stylegan} and ours fail to reconstruct the input images.
\Tbl{\ref{tbl:quantitative_results_on_ffhq}} reports a quantitative comparison in PSNR and FID~\cite{FID}.
We refer the readers to our supplementary material for additional comparison in SSIM~\cite{ssim} and RMSE.
The table shows that our $16\times16$ version achieves the highest reconstruction quality both in PSNR and FID for all geometric transformations.
Both in the figure and table, Im2StyleGAN shows high-quality reconstruction results.
However, due to the lack of in-domain constraints, Im2StyleGAN tends to produce out-of-domain latent codes that are not semantically editable as will be seen later in this section.
The table also shows that the performances of the previous methods degrade quickly for larger translations and rotations.
For example, the performance of P-norm$^+$~\cite{pnorm} drops by 3.86 dB for the rotation by 30 degrees.
Our $8\times8$ version performs worse than the $16\times16$ version as it uses a more constrained latent space.
We also note that our $16\times16$ version outperforms all the other methods even for images without geometric transformations (Translation = 0 in \Tbl{\ref{tbl:quantitative_results_on_ffhq}}) thanks to the base code $\mathbf{f}$ supporting local variations.

\paragraph{Inversion of natural images}
Due to the large diversity of natural images, it is difficult to accurately reconstruct and edit a natural image using previous GAN inversion approaches.
On the other hand, thanks to the high degree-of-freedom of the $\mathcal{F}/\mathcal{W}^+$ space, our approach is especially effective in handling natural images.
To verify this, we compare the reconstruction and editing quality of previous methods and ours on natural images.
For evaluation, we use StyleGAN and StyleGAN2 models~\cite{stylegan,stylegan2} pre-trained on the LSUN bedroom, tower and cat datasets~\cite{lsun}.
We also collected 25 bedroom, tower and cat images each from the internet and used them as our test sets
so that the images in the test sets are of the same classes as the training images of the pre-trained models, but not aligned with the training images.
For these datasets, we use 3,000 iterations.

We compare our method against Im2StyleGAN~\cite{im2stylegan}, which shows high-quality reconstruction results in the previous experiment, and IDinvert~\cite{indomain_gan}, which finds an in-domain latent code for semantic editing.
We use the authors' code for IDinvert.
\Fig{\ref{fig:general_results}} shows a qualitative comparison of the reconstruction and editing qualities.
Both Im2StyleGAN~\cite{im2stylegan} and IDinvert~\cite{indomain_gan} produce less accurate reconstruction results than ours.
Their editing results also show artifacts due to the out-of-range input images.
Especially, the editing results of Im2StyleGAN have severe artifacts as its out-of-domain latent codes.
In contrast, our method shows high-quality reconstruction and editing results for all three cases.
\Tbl{\ref{tbl:quantitative_results_on_general_datasets}} shows a quantitative comparison of the reconstruction qualities.
The table also shows that our method achieves high reconstruction quality on natural images compared to the other methods.
More results can be found in the supplementary material.

\begin{figure}[t]
\centering
\includegraphics[width=0.91\linewidth]{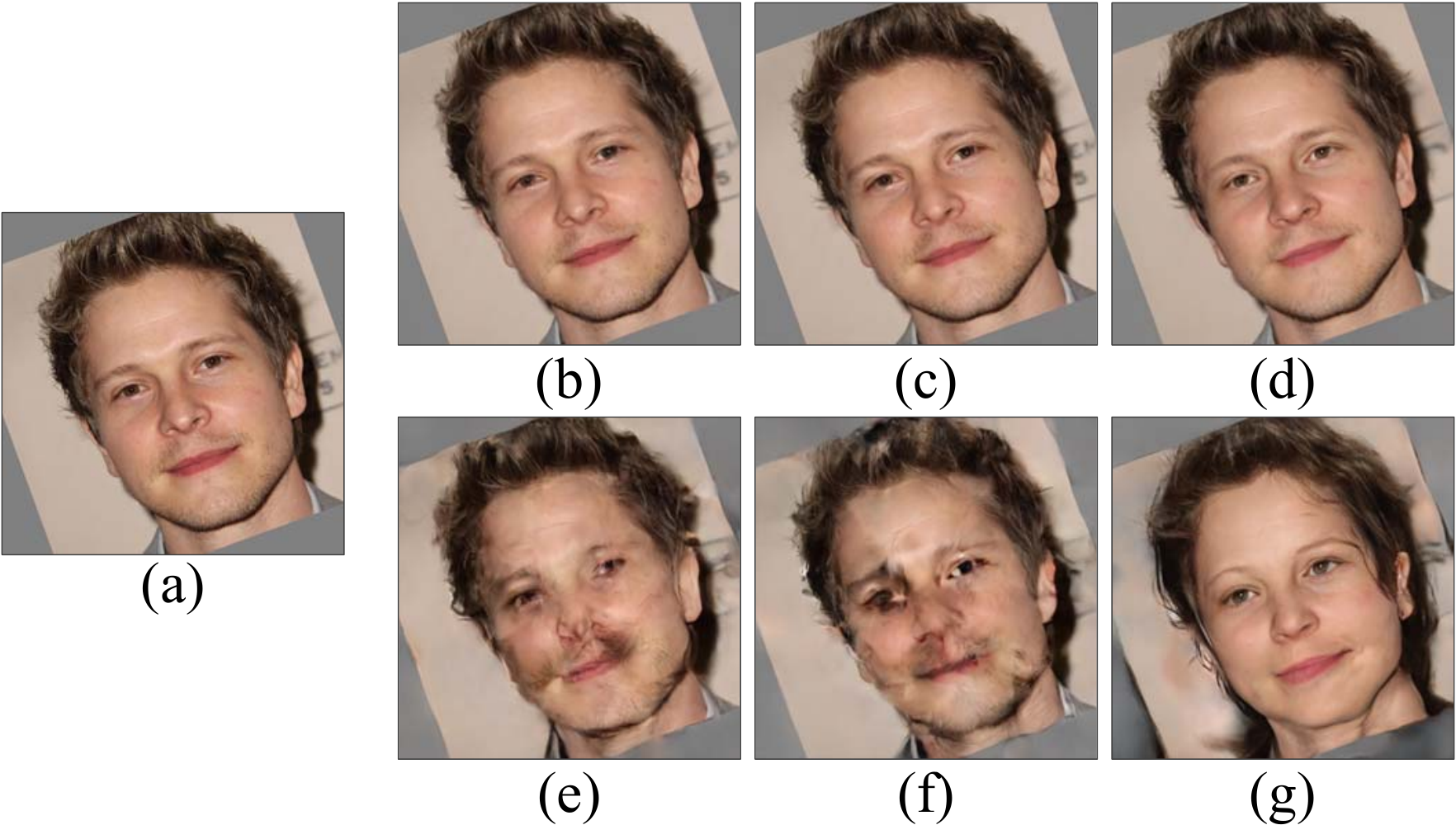}
\vspace{-0.2cm}
\caption{Ablation study. (a) Target image. (b) Reconstruction using only the reconstruction loss. (c) Reconstruction using the regularization on $\textbf{w}_{M+}$. (d) Reconstruction using the regularization both on $\textbf{w}_{M+}$ and $\textbf{f}$ (our final method). Semantic editing results of (b), (c) and (d) are shown in (e), (f) and (g), respectively.}
\label{fig:ablation_study}
\vspace{-0.3cm}
\end{figure}

\paragraph{Ablation study}
\Fig{\ref{fig:ablation_study}} shows a qualitative comparison of variants of our method using StyleGAN2~\cite{stylegan2} to verify the effectiveness of our regularization scheme.
While all the variants show excellent reconstruction results thanks to the high degree of freedom of the $\mathcal{F}/\mathcal{W}^+$ space,
the editing results of the variants that use only the reconstruction loss or regularization on the detail code $\textbf{w}_{M+}$ are severely degraded.
On the other hand, the editing result of our final model in (d) looks the most natural thanks to our regularized inversion scheme.
More examples and a quantitative evaluation are in the supplementary material.

\begin{figure}[t]
\centering
\includegraphics[width=0.91\linewidth]{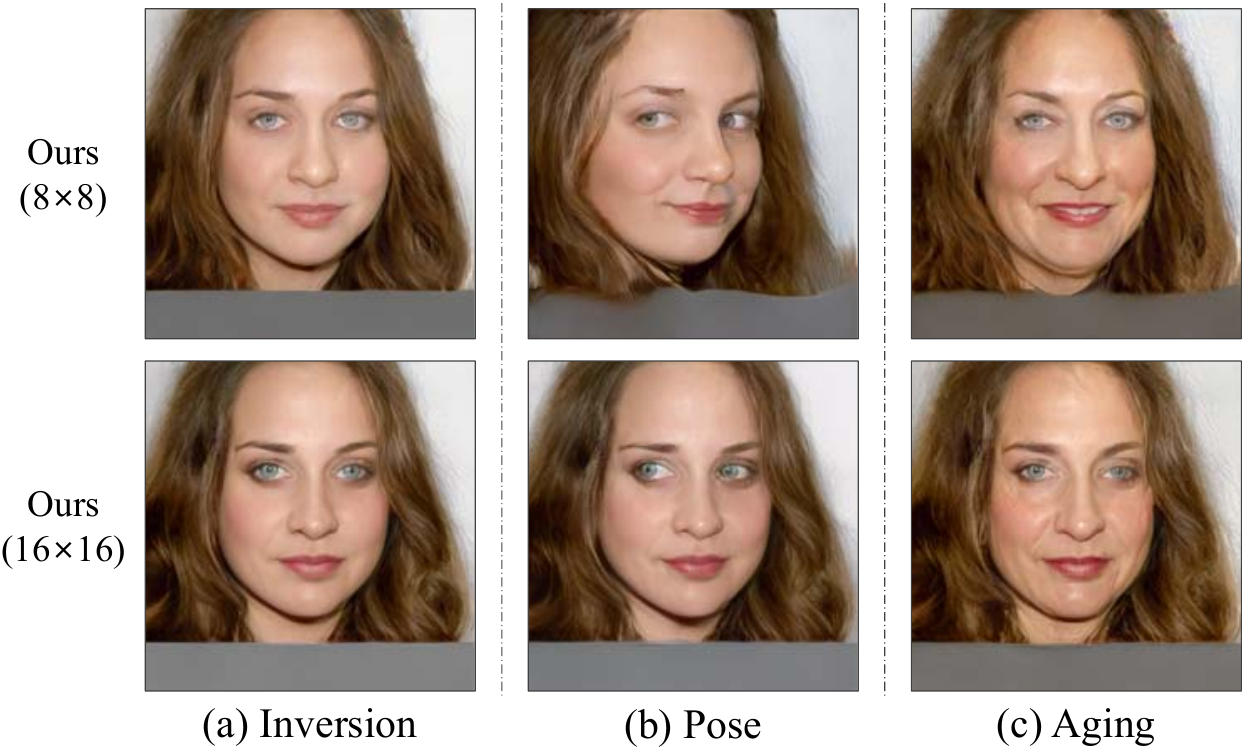}
\vspace{-0.2cm}
\caption{Available editing operations according to the size of $\mathbf{f}$. (a) shows inversion results using $\mathbf{f}$ of different sizes. (b) and (c) show results of different editing operations. The pose editing operation that changes the overall image structure does not work for $\mathbf{f}$ of size $16\times16$ while the aging operation works for both.}
\label{fig:analysis_by_layer}
\vspace{-0.3cm}
\end{figure}

\paragraph{Editing operations \textit{v.s.} scale of base code $\mathbf{f}$}
Finally, we analyze the effect of the scale of the base code $\mathbf{f}$ on image editing.
Using a feature map at a finer-scale for the base code $\mathbf{f}$ leads to higher reconstruction quality as shown in \Fig{\ref{fig:ffhq_results}} and \Tbl{\ref{tbl:quantitative_results_on_ffhq}}.
On the other hand, it also reduces the diversity of semantic editing operations.
Especially, it makes it difficult to perform semantic operations that rely on coarse-scale latent codes $\mathbf{w}_i$ in the $\mathcal{W}$ space.
\Fig{\ref{fig:analysis_by_layer}} shows an example.
While our method with $\mathbf{f}$ of size $8\times8$ supports both pose changing and aging,
ours with $\mathbf{f}$ of size $16\times16$ does not support pose changing since the pose changing operation requires to edit small-scale latent codes.

\section{Conclusion}
\label{sec:conclusion}
In this paper, we proposed \emph{BDInvert}, a novel GAN inversion approach for semantic editing of out-of-range images with geometric transformations.
Based on the StyleGAN and StyleGAN2 frameworks~\cite{stylegan,stylegan2}, we presented an alternative latent space $\mathcal{F}/\mathcal{W}^+$ that supports geometric transformations of an image as well as its semantic manipulation.
To find a proper solution in the $\mathcal{F}/\mathcal{W}^+$ space that is semantically editable, we introduced a novel regularized optimization approach.
We verified the effectiveness of our approach both qualitatively and quantitatively.

\paragraph{Limitations}
As discussed in \Secs{\ref{sec:latent_space}} and \ref{sec:experiments}, the $\mathcal{F}/\mathcal{W}^+$ space reduces the diversity of semantic editing operations.
Also as our approach is based on optimization, it requires a relatively long computation time.
With an Nvidia RTX 3090 GPU, it takes about 3 minutes for a $1024\times1024$-sized image.
Our approach cannot handle images with severe geometric transformations.
However, this can be easily resolved by rough alignment of an input image as our method does not require accurate alignment.
Finally, our method cannot handle images that are too different from the training dataset.
See the supplementary material for examples.

\paragraph{\small Acknowledgements}
{\small This work was supported by Institute of Information \& communications Technology Planning \& Evaluation (IITP) grant funded by the Korea government(MSIT) (No.2019-0-01906, Artificial Intelligence Graduate School Program(POSTECH)) and National Research Foundation of Korea (NRF) grant funded by the Korea government(MSIT) (NRF-2018R1A5A1060031, No. 2020R1C1C1014863).}

\end{document}